\documentclass[a4paper]{article} %

\usepackage{graphicx}
\usepackage{booktabs}
\usepackage{algorithm}
\usepackage{algpseudocode}
\usepackage{xspace}
\usepackage{soul}

\usepackage{amsmath}
\usepackage{amssymb}
\usepackage{booktabs}       %
\usepackage{hyperref}

\usepackage{blindtext,titlefoot}
\usepackage{float}

\newcommand{\algoname}{RNet--DQN\xspace}
\newcommand{\mdpshort}{GC-MDP\xspace}
\newcommand{\mdplong}{Graph Construction MDP\xspace}
\newcommand{\mdpfull}{Graph Construction Markov Decision Process\xspace}

\newcommand{\mathcolorbox}[2]{\colorbox{#1}{$\displaystyle #2$}}

\newcommand\showhighlights{0}

\newcommand{\customhl}[1]{%
\ifnum1=\showhighlights \relax
  \hl{#1}
\else
  #1
\fi
}

\newcommand{\customhlmath}[1]{%
\ifnum1=\showhighlights \relax
  \mathcolorbox{yellow}{#1}
\else
  #1
\fi
}

\DeclareMathOperator*{\argmax}{\arg\!\max}
\DeclareMathOperator*{\argmin}{\arg\!\min}

\begin{document}

\title{Goal-directed graph construction using reinforcement learning}%

\author{%
Victor-Alexandru Darvariu$^{1,2}$, Stephen Hailes$^{1}$ and Mirco Musolesi$^{1,2,3}$\\
\small{$^{1}$Department of Computer Science, University College London, London, UK}\\
\small{$^{2}$The Alan Turing Institute, London, UK}\\
\small{$^{3}$Department of Computer Science and Engineering, University of Bologna, Bologna, Italy}\\
\small{\texttt{\{v.darvariu,\enspace s.hailes,\enspace m.musolesi@ucl.ac.uk\}}}
}

\date{\vspace{-5ex}}
\maketitle
\unmarkedfntext{This article has been published in \textit{Proceedings of the Royal Society A} \textbf{477}:20210168. \url{http://doi.org/10.1098/rspa.2021.0168}. A previous version of this article was titled ``Improving the Robustness of Graphs through Reinforcement Learning and Graph Neural Networks".}

\begin{abstract}
Graphs can be used to represent and reason about systems and a variety of metrics have been devised to quantify their global characteristics. However, little is currently known about how to construct a graph or improve an existing one given a target objective. In this work, we formulate the construction of a graph as a decision-making process in which a central agent creates topologies by trial and error and receives rewards proportional to the value of the target objective. By means of this conceptual framework, we propose an algorithm based on reinforcement learning and graph neural networks to learn graph construction and improvement strategies. Our core case study focuses on robustness to failures and attacks, a property relevant for the infrastructure and communication networks that power modern society. Experiments on synthetic and real-world graphs show that this approach can outperform existing methods while being cheaper to evaluate. It also allows generalization to out-of-sample graphs, as well as to larger out-of-distribution graphs in some cases. The approach is applicable to the optimization of other global structural properties of graphs.
\end{abstract}

\section{Introduction}\label{intro}

Graphs are mathematical abstractions that can be used to model a variety of systems, from infrastructure and biological networks to social structures. Various methods for analysing networks have been developed: these have been often used for understanding the systems themselves and range from mathematical models of how families of graphs are generated~\cite{watts_collective_1998,barabasi_emergence_1999} to measures of centrality for capturing the roles of vertices~\cite{bianchini_inside_2005} and global network characteristics~\cite{newman_networks_2018}, to name but a few.

A measure that has attracted significant interest from researchers and practitioners is \textit{robustness}~\cite{newman_structure_2003} (sometimes called \textit{resilience}), which is typically defined as the capacity of the graph to withstand random failures, targeted attacks on key nodes, or some combination thereof. A network is considered robust if a large fraction (\textit{critical fraction}) of nodes have to be removed before it becomes disconnected~\cite{cohen_resilience_2000}, its diameter increases \cite{albert_error_2000}, or its largest connected component diminishes in size \cite{beygelzimer_improving_2005}. Previous work has focused on the robustness of communication networks such as the Internet~\cite{cohen_breakdown_2001} and infrastructure networks used for transportation and energy distribution~\cite{cetinay_nodal_2018}, for which resilience is a key property.

In many practical cases, an initial network is given and the only way of improving its robustness is through the modification of its structure.
This problem was first approached by considering edge addition or rewiring, based on random and preferential (w.r.t. node degree) modifications~\cite{beygelzimer_improving_2005}. Alternatively, a strategy has been proposed that uses a ``greedy" modification scheme based on random edge selection and swapping if the resilience metric improves~\cite{schneider_mitigation_2011}. Another line of work focuses on the spectral decomposition of the graph Laplacian, and using properties such as the algebraic connectivity~\cite{wangAlgebraicConnectivityOptimization2008} and effective graph resistance~\cite{wang_improving_2014} to guide modifications. While simple and interpretable, these strategies may not yield the best solutions or generalise across networks with varying characteristics and sizes. Certainly, better solutions may be found by exhaustive search, but the time complexity of exploring all the possible topologies and the cost of computing the metric render this strategy infeasible. With the goal of discovering better strategies than existing methods, we ask whether \textit{generalisable network construction strategies for improving robustness can be learned}.

\begin{figure*}[t]
  \includegraphics[width=\textwidth,trim={2.25mm 3.5mm 0 2.75mm},clip=true]{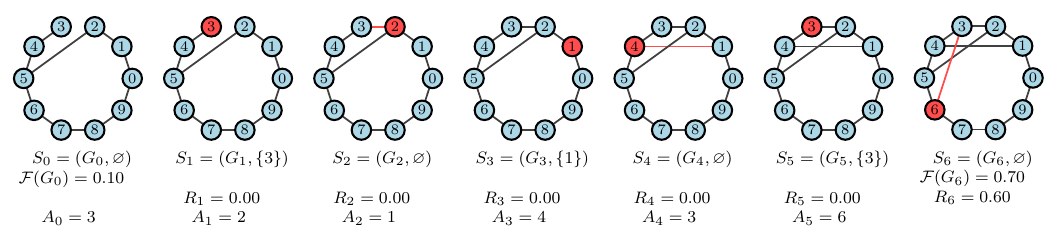}
     \caption{Illustration of a \mdplong (\mdpshort) trajectory. The agent is provided with a start state $S_0=(G_0,\varnothing)$. It must make $L=3$ edge additions over a sequence of 6 node selections (actions $A_t$), receiving rewards $R_t$ proportional to the value of an objective function $\mathcal{F}$ applied to the graph. In this case, $\mathcal{F}$ quantifies the robustness of the network to targeted node removal, computed by removing nodes in decreasing order of their degree and in decreasing order of the labels if two nodes have the same degree. We observe an improvement of the robustness of the graph from $\mathcal{F}(G_0)=0.1$ to $\mathcal{F}(G_6)=0.7$. Actions and the corresponding edges are highlighted.} 
  \label{gno_illustration}
\end{figure*}

Starting from this motivation, we formalise the process of graph construction and improvement as a Markov Decision Process (MDP) in which rewards are proportional to the value of a graph-level objective function. We consider two objective functions that quantify robustness as the critical fraction of the network in the presence of random failures and targeted attacks. Inspired by recent successes of RL in solving combinatorial optimisation problems on graphs~\cite{bello_neural_2016,khalil_learning_2017}, we make use of Graph Neural Network (GNN) architectures~\cite{gilmer_neural_2017} together with the Deep Q-Network (DQN)~\cite{mnih_human_2015} algorithm. Recent work in goal-directed graph generation and improvement considers performing edge additions for adversarially attacking GNN classifiers~\cite{dai_adversarial_2018} and generating molecules with certain desirable properties using domain-specific rewards~\cite{you_graph_2018}. In contrast, to the best of our knowledge, this is the first time that RL is used \textit{to learn how to construct a graph such as to optimise a global structural property}. While in this paper we focus on robustness, other intrinsic global properties of graphs, such as efficiency~\cite{latoraEfficientBehaviorSmallWorld2001} or communicability~\cite{estrada_communicability_2008}, could be used as optimisation targets.

The contribution of this paper is twofold. Firstly, we propose a framework for improving global structural properties of graphs, by introducing the \mdpfull (\mdpshort). Secondly, focusing on the robustness of graphs under failures and attacks as a core case study, we offer an in-depth empirical evaluation that demonstrates significant advantages over existing approaches in this domain, both in terms of the quality of the solutions found as well as the time complexity of model evaluation. Since this approach addresses the problem of building robust networks with a DQN, we name it \textit{\algoname}.

The remainder of the paper is structured as follows. We provide the definitions of the \mdpshort and the robustness measures in Section~\ref{problem}. Section~\ref{fnapprox} describes state and action representations for deep RL using GNNs. We present our experimental setup in Section~\ref{experimental}, and discuss our main results in Section~\ref{results}. In Section~\ref{related} we review and compare the key works in this area. Finally, we conclude and offer a discussion of avenues for future work in Section~\ref{conclusion}.
\section{Robust Graph Construction as a Decision Making Problem}\label{problem}
\paragraph{MDP Preliminaries.} An MDP is one possible formalisation of a decision making process. The decision maker, called an \textit{agent}, interacts with an \textit{environment}.  When in a \textit{state} $s\in \mathcal{S}$, the agent must take an \textit{action} $a$ out of the set $\mathcal{A}(s)$ of valid ones, receiving a \textit{reward} $r$ governed by the reward function $\mathcal{R}(s,a)$. Finally, the agent finds itself in a new state $s'$, depending on a transition model $\mathcal{P}$ that governs the joint probability distribution $P(s',a,s)$ of transitioning to state $s'$ after taking action $a$ in state $s$. This sequence of interactions gives rise to a \textit{trajectory}.
The agent's goal is to maximise the expected (possibly discounted) sum of rewards it receives over all the trajectories. The tuple $(\mathcal{S,A,P,R, \gamma})$ defines this MDP, where $\gamma \in [0,1]$ is a discount factor. We also define a \textit{policy} $\pi(a|s)$, i.e., a distribution of actions over states, which fully determines the behaviour of the agent. Given a policy $\pi$, the \textit{action-value function} $Q_{\pi}(s,a)$ is defined as the expected return when starting from $s$, taking action $a$, and subsequently following policy $\pi$. 

\paragraph{Modelling Graph Construction.} Let $\mathbf{G}^{(N)}$ be the set of labelled, undirected, unweighted graphs with $N$ nodes; each such graph $G=(V,E)$ consists of a vertex set $V$ and edge set $E$. Let $\mathbf{G}^{(N,m)}$ be the subset of $\mathbf{G}^{(N)}$ with $|E|=m$. We also let $\mathcal{F}\colon \mathbf{G}^{(N)} \to [0,1]$ be an objective function, and $L\in\mathbb{N}$ be a modification budget. Given an initial graph $G_0=(V,E_0)\in\mathbf{G}^{(N,m_0)}$, the aim is to perform a series of $L$ edge additions to $G_0$ such that the resulting graph $G_*=(V,E_*)$ satisfies: 

\begin{align*}
    G_*&=\argmax_{G' \in \mathbf{G'}} {\mathcal{F}(G')},\\
    \text{where}\quad \mathbf{G}' = \{ G&=(V,E) \in \mathbf{G}^{(N, m_0+L)}\ |\ E_0 \subset E\}.
\end{align*}

This combinatorial optimisation problem can be cast as a sequential decision-making process. In order to enable scaling to large graphs, the agent has to select a node at each step, and an edge is added to the graph after every two decisions~\cite{dai_adversarial_2018}. Tasks are episodic; each episode proceeds for at most $2L$ steps. A trajectory visualisation is shown in Figure~\ref{gno_illustration}. Formally, we define the \mdplong (\mdpshort) as follows: 

\begin{enumerate}
\item \textit{State}: The state $S_t$ is a tuple $\left(G_t,\sigma_t\right)$ containing the graph $G_t=(V, E_t)$ and an \textit{edge stub} $\sigma_t$. $\sigma_t$ can be either the empty set $\varnothing$ or the singleton $\{v\}$, where $v \in V$.
\item \textit{Action}: $A_t$ corresponds to the selection of a node in $V$. Letting the degree of node $v$ be $d_v$, available actions are defined as:
\begin{align*}
\mathcal{A}(S_t=((V, E_t), \varnothing)) &= \{v \in V\ |\ d_v < |V|-1 \} \\
\mathcal{A}(S_t=((V, E_t), \{ \sigma_t \})) &= \{v \in V\ |\ (\sigma_t, v) \notin E_t \}
\end{align*}

\item \textit{Transitions}: The transition model is defined as $P(S_t=s'|S_{t-1}=s,A_{t-1}=a) = \delta_{S_tS'}$,
\[
\text{where}\ S'=
  \begin{cases}
    \left(\left(V,E_{t-1} \cup \left(\sigma_{t-1}, a \right) \right), \varnothing \right),\  \text{if}\  2 \mid t \\
    \left(\left(V,E_{t-1}\right), \{a\} \right),\  \text{otherwise}
  \end{cases}
\]
\item \textit{Reward}: The reward $R_t$ is defined as follows\footnote{Since $\mathcal{F}$ is very expensive to estimate, we deliberately only provide the reward at the end of the episode in order to make the training feasible computationally, to the detriment of possible credit assignment issues. Intermediate rewards based on the true objective or a related quantity represent a middle ground which we leave for future work.}:
\[
R_t =
  \begin{cases}
    \mathcal{F}(G_{t}) - \mathcal{F}(G_0),\  \text{if}\   t=2L \\
    0,\  \text{otherwise}
  \end{cases}
\]
\end{enumerate}

\paragraph{Definition of Objective Functions for Robustness.} We are interested in the robustness of graphs as objective functions. Given a graph $G$, we let the \textit{critical fraction} $p(G, \xi) \in [0,1]$ be the minimum fraction of nodes that have to be removed from $G$ in some order $\xi$ for it to become disconnected (i.e., have more than one connected component). Connectedness is a crucial operational constraint and the higher this fraction is, the more robust the graph can be said to be.
\footnote{We note that while connectedness is required for the specific objective functions considered in this work, it is not required by either the \mdpshort formulation or the learning mechanism itself. Other robustness objectives that quantify, e.g., the size of the largest connected component are also applicable, as are fundamentally different objectives.}
The order $\xi$ in which nodes are removed can have an impact on $p$, \customhl{and corresponds to different scenarios: random removal is typically used to model arbitrary failures, while targeted removal is adopted as a model for attack}. Formally, we consider both random permutations $\xi_{random}$ of nodes in $G$, as well as permutations $\xi_{targeted}$, which are subject to the constraint that nodes must appear in the order of their degree, i.e.,
\[
\forall v,u \in V .\ \xi_{targeted}(v) \leq \xi_{targeted}(u) \iff d_v \geq d_u
\]
We define the objective functions $\mathcal{F}$ in the following way:

\begin{enumerate}
\item \textit{Expected Critical Fraction to Random Removal}: 
\[
 \mathcal{F}_{random}(G) = \mathbb{E}_{\xi_{random}}[p(G, \xi_{random})]
\]
\item \textit{Expected Critical Fraction to Targeted Removal}: 
\[
 \mathcal{F}_{targeted}(G) = \mathbb{E}_{\xi_{targeted}}[p(G, \xi_{targeted})]
\]
\end{enumerate}

We use Monte Carlo (MC) sampling for estimating these quantities. For completeness, Algorithm 1 in the Supplementary Material describes how the simulations are performed. In the remainder of the paper, we use $\mathcal{F}_{random}(G)$ and $\mathcal{F}_{targeted}(G)$ to indicate their estimates obtained in this way. We highlight that evaluating an MC sample has time complexity $O(|V| \times (|V|+|E|))$: it involves checking connectedness (an $O(|V|+|E|)$ operation) after the removal of each of the $O(|V|)$ nodes. Typically, many such samples need to be used to obtain a low-variance estimate of the quantities. Coupled with the number of possible topologies, the high cost renders even shallow search methods infeasible in this domain.

\section{Learning to Build Robust Graphs with Function Approximation}\label{fnapprox}

While the problem formulation described in Section~\ref{problem} may allow us to work with a tabular RL method, the number of states quickly becomes intractable -- for example, there are approximately $10^{57}$ labelled, connected graphs with 20 vertices~\cite{oeisNumGraphs2020}. Thus, we require a means of considering graph properties that are label-agnostic, permutation-invariant, and generalise across similar states and actions. Graph Neural Network architectures address these requirements. In particular, we use a graph representation based on a variant of structure2vec (S2V) \cite{dai_discriminative_2016}, a GNN architecture inspired by mean field inference in graphical models.~\footnote{The problem formulation does not depend on the specific GNN or RL algorithm used. While further advances developed by the community in these areas~\cite{maronProvablyPowerfulGraph2019,hesselRainbowCombiningImprovements2018} can be incorporated, in this paper we focus on aspects specific to the challenges of optimising the global properties of graphs.}
Given an input graph $G=(V,E)$ where nodes $v \in V$ have feature vectors $\mathbf{x}_v$, its objective is to produce for each node $v$ an embedding vector $\mu_v$ that captures the structure of the graph as well as interactions between neighbours. This is performed in several rounds of aggregating the features of neighbours and applying an element-wise non-linear activation function such as the rectified linear unit. For each round $k \in \{1,2,...,K\}$, the network simultaneously applies updates of the form:
\[
\customhlmath{\mu_v^{(k+1)} = \text{relu}\bigg( \theta^{(1)}\mathbf{x}_v + \ \theta^{(2)} \sum_{u \in \mathcal{N}(v)}{\mu_u^{(k)}}\bigg)}
\]
where $\mathcal{N}(v)$ is the neighbourhood of node $v$. We initialise embeddings with $\mu_v^{(0)}=\mathbf{0}\ \forall v \in V$, and let $\mu_v=\mu_v^{(K)}$. Once node-level embeddings are obtained, permutation-invariant embeddings for a subgraph $\mathcal{S}$ can be derived by summing the node-level embeddings: $\mu(\mathcal{S}) = \sum_{v_i \in \mathcal{S}}{\mu_{v_i}}$. The node features $\mathbf{x}_v$ are one-hot 2-dimensional vectors representing whether $v$ is the edge stub, and their use is required to satisfy the Markovian assumption behind the MDP framework (the agent ``commits" to selecting $v$, which is now part of its present state). Each state contains at most one edge stub.
 
In Q-learning~\cite{watkins_q_1992}, the agent estimates the action-value function $Q(s,a)$ introduced earlier, and derives a deterministic policy that acts greedily with respect to it. The agent interacts with the environment and updates its estimates according to:
\[
Q(s,a) \leftarrow Q(s,a) + \alpha [r + \gamma \max_{a' \in \mathcal{A}(s')}{Q(s',a') - Q(s,a)} ]
\]
During learning, exploratory random actions are taken with probability $\epsilon$. In the case of high-dimensional state and action spaces, approaches that use a neural network to estimate $Q(s,a)$ have been successful in a variety of domains ranging from general game-playing to continuous control~\cite{mnih_human_2015,lillicrap_continous_2015}. In particular, we use the DQN algorithm: a sample-efficient method that improves on neural fitted Q-iteration~\cite{riedmiller_neural_2005} by use of an experience replay buffer and an iteratively updated target network for action-value function estimation. Specifically, we use two parametrisations of the Q-function depending on whether the state $S_t$ contains an edge stub:
\begin{align*}
Q(S_t=(G_t, \varnothing), A_t) &= \theta^{(3)} \text{relu}\ (\theta^{(4)} \left[   \mu_{A_t}, \mu({G_t})    \right] )  \\
Q(S_t=(G_t, \{ \sigma_t \} ), A_t) &= \theta^{(5)} \text{relu}\ (\theta^{(6)} \left[   \mu_{\sigma_t}, \mu_{A_t}, \mu({G_t})    \right] )
\end{align*}

where $\left[\ \cdot, \cdot    \right]$ represents concatenation. This lets the model learn \textit{combinations} of relevant node features (e.g., that connecting two central nodes has high Q-value). The use of GNNs has several advantages: firstly, the parameters $\Theta=\{\theta^{(i)}\}_{i=1}^{6}$ can be learned in a goal-directed fashion for the RL objective, allowing for flexibility in the learned representation. Secondly, the embeddings have the potential to generalise to larger graphs since they control how to combine node features of neighbours in the message passing rounds and are not restricted to graphs of a particular size. We note that the underlying S2V parameters $\theta^{(1)}, \theta^{(2)}$ are shared between the two Q-function parametrisations.

\section{Experimental Setup}\label{experimental}
\paragraph{Learning Environment.} We build a learning environment that allows for the definition of an arbitrary graph objective function $\mathcal{F}$ and provides a standardised interface for agents. Our implementation of the environment, \algoname and baseline agents, and experimental suite is provided as a code repository containing Docker image blueprints that enable the reproduction of the results presented herein (up to hardware differences), including the relevant tables and figures. \customhl{The instructions about how to obtain, configure, and run the code are provided in the Supplementary Material.}

\paragraph{Baselines.} We compare against the following approaches:
\begin{itemize}
\item \textit{Random}: This strategy randomly selects an available action.
\item \textit{Greedy}: This strategy uses lookahead and selects the action that gives the biggest improvement in the estimated value of $\mathcal{F}$ over one edge addition.
\item \textit{Preferential}: Previous works have considered preferential additions between nodes with the two lowest degrees~\cite{beygelzimer_improving_2005}, connecting a node with the lowest degree to a random node~\cite{wangAlgebraicConnectivityOptimization2008} or connecting the two nodes with the lowest degree product~\cite{wang_improving_2014}, i.e., adding an edge between the vertices $v,u$ that satisfy $\argmin_{v,u} {d_v \cdot d_u}$. We find the latter works best in all settings tested, and refer to it as \textit{LDP}.
\item \textit{Fiedler Vector (FV)}: The concept of Fiedler Vector was introduced by \cite{fiedler1973algebraic} and for robustness improvement by~\cite{wangAlgebraicConnectivityOptimization2008}. This strategy adds an edge between the vertices $v,u$ that satisfy $\argmax_{v,u} {|\mathbf{y}_v - \mathbf{y}_u|}$, where $\mathbf{y}$ is the Fiedler vector i.e., the eigenvector of the graph Laplacian $\mathcal{L}$ corresponding to the second smallest eigenvalue.
\item \textit{Effective Resistance (ERes)}: The concept of effective resistance was introduced by~\cite{ellens_effectivegraph_2011} and for robustness improvement as a local pairwise approximation by~\cite{wang_improving_2014}. This strategy selects vertices $v,u$ that satisfy $\argmax_{v,u} {\Omega_{v,u}}$. $\Omega_{v,u}$ is defined as $(\hat{\mathcal{L}}^{-1})_{vv} + (\hat{\mathcal{L}}^{-1})_{uu} -2(\hat{\mathcal{L}}^{-1})_{vu}$, where $\hat{\mathcal{L}}^{-1}$ is the pseudoinverse of $\mathcal{L}$.
\item \textit{Supervised Learning (SL)}: We consider a supervised learning baseline by regressing on $\mathcal{F}$ to learn an approximate $\hat{\mathcal{F}}$. We use the same S2V architecture as \algoname, which we train using MSE loss instead of the Q-learning loss. To select actions for a graph $G$, the agent considers all graphs $G'$ that are one edge away, selecting the one that satisfies $\argmax_{G'} {\hat{\mathcal{F}}}$.
\end{itemize}

\paragraph{Neural Network Architecture.} For all experiments, we use an S2V embedding vector of length $64$. The neural network architecture used for \algoname and SL is formed of state-action embeddings obtained using S2V followed by a multi-layer perceptron; the single output unit corresponds to the $Q(s,a)$ estimate for \algoname and the predicted $\hat{\mathcal{F}}$ for SL respectively. Details of hyperparameters used for the two learning-based models are provided in the Supplementary Material.

\paragraph{Evaluation Protocol.} We evaluate \algoname and baselines both on synthetic and real-world graphs. We allow agents a number of edge additions $L$ equivalent to a percentage $\tau$ of total possible edges. As an evaluation metric, we report the cumulative reward obtained by the agents, which quantifies the improvement in the objective function value between the final and original graphs. Specifically, in the context of the robustness metrics used, the values provided measure the difference in the expected fraction of nodes that need to be removed for the network to become disconnected. Training is performed separately for each graph family, objective function $\mathcal{F}$, and value of $L$. Where an agent is non-deterministic (either through intrinsic stochasticity or need for training), we repeat its evaluation (and training where applicable, starting from a different random initialisation of the network weights) to compute confidence intervals. For the learned models, we record both average and maximum performance. No hyperparameter tuning is performed due to computational budget constraints. Details about the experimental settings are provided in the Supplementary Material.

\paragraph{Synthetic Graphs.} We consider graphs generated through the following models:
\begin{itemize}
\item \textit{Erd\H{o}s--R\'{e}nyi (ER)}: A graph sampled uniformly out of $\mathbf{G}^{(N,m)}$~\cite{erdos_evolution_1960}. We use $m =\frac{20}{100} * \frac{N * (N-1)}{2}$, which represents 20\% of all possible edges.
\item \textit{Barab\'{a}si--Albert (BA)}: A growth model where $n$ nodes each attach preferentially to $M$ existing nodes~\cite{barabasi_emergence_1999}. We use $M=2$.
\end{itemize}

We consider graphs with $|V|=20$, allowing agents to add a percentage of the total number of edges equal to $\tau \in \{1,2,5\}$, which yields $L\in \{2,5,10\}$. For \algoname and SL, we train on a disjoint set of graphs $\mathbf{G}^{train}$. We periodically measure performance on another set $\mathbf{G}^{validate}$, storing the best model found. We use $|\mathbf{G}^{train}| = 10^4$ and $|\mathbf{G}^{validate}| = 10^2$. The performance of all agents is evaluated on a set $\mathbf{G}^{test}$ with $|\mathbf{G}^{test}| = 10^2$ generated using the ER and BA models. In order to evaluate out-of-distribution generalisation, we repeat the evaluation on graphs with up to $|V|=100$ (only up to $|V|=50$ for Greedy and SL due to computational cost, see next section) and scale $m$ (for ER) and $L$ accordingly. For non-deterministic agents, evaluation (and training, where applicable) is repeated across $50$ random seeds.

\paragraph{Real-World Graphs.} In order to evaluate our approach on real-world graphs, we consider infrastructure networks (for which robustness is a critical property) extracted from two datasets: \textit{Euroroad} (road connections in mainland Europe and parts of Western and Central Asia~\cite{subeljRobustNetworkCommunity2011,kunegisKONECTKoblenzNetwork2013}, $|V|=1174$) and \textit{Scigrid} (a dataset of the European power grid~\cite{medjroubiOpenDataPower2017}, $|V|=1479$). We split these graphs by the country in which the nodes are located, selecting the largest connected component in case they are disconnected. We then select those with $20 \leq |V| \leq 50$, obtaining $6$ infrastructure graphs for Scigrid and $8$ for Euroroad. The partitioning and selection procedure yields infrastructure graphs for the following countries:
\begin{itemize}
    \item \textit{Euroroad}: Finland, France, Kazakhstan, Poland, Romania, Russia, Turkey, Ukraine.
    \item \textit{Scigrid}: Switzerland, Czech Republic, United Kingdom, Hungary, Ireland, Sweden.
\end{itemize}

Since, in this context, the performance on individual instances matters more than generalisability, we train and evaluate on each graph separately (effectively, the sets $\mathbf{G}^{train},\mathbf{G}^{validate},\mathbf{G}^{test}$ all have cardinality $1$ and contain the same graph). \textit{SL} is excluded for this experiment since we consider a single network. Evaluation is repeated across $10$ random seeds.

\section{Results}\label{results}

\begin{table*}
\centering
\begin{small}
\resizebox{\columnwidth}{!}{
\begin{tabular}{ccc|ccccccccc}
                 &                &  &       Random &   LDP &    FV &  ERes &  Greedy &           \multicolumn{2}{c}{SL}   &     \multicolumn{2}{c}{\algoname}   \\
Objective & $\mathbf{G}$ & L &              &       &       &       &         & \textit{avg}& \textit{best}& \textit{avg}& \textit{best} \\
\midrule
$\mathcal{F}_{random}$ & BA & 2  &  0.018\tiny{$\pm0.001$} & 0.036 & 0.051 & 0.053 &   0.033 &  0.048\tiny{$\pm0.002$} &    \textbf{0.057} &  0.051\tiny{$\pm0.001$} &          \textbf{0.057} \\
                 &                & 5  &  0.049\tiny{$\pm0.002$} & 0.089 & 0.098 & 0.106 &   0.079 &  0.099\tiny{$\pm0.003$} &    0.122 &  0.124\tiny{$\pm0.001$} &          \textbf{0.130} \\
                 &                & 10 &  0.100\tiny{$\pm0.003$} & 0.158 & 0.176 & 0.180 &   0.141 &  0.161\tiny{$\pm0.008$} &    0.203 &  0.211\tiny{$\pm0.001$} &          \textbf{0.222} \\
                 & ER & 2  &  0.029\tiny{$\pm0.001$} & 0.100 & 0.103 & 0.103 &   0.082 &  0.094\tiny{$\pm0.001$} &    0.100 &  0.098\tiny{$\pm0.001$} &          \textbf{0.104} \\
                 &                & 5  &  0.071\tiny{$\pm0.002$} & 0.168 & 0.172 & \textbf{0.175} &   0.138 &  0.158\tiny{$\pm0.002$} &    0.168 &  0.164\tiny{$\pm0.001$} &          0.173 \\
                 &                & 10 &  0.138\tiny{$\pm0.002$} & 0.238 & 0.252 & \textbf{0.253} &   0.217 &  0.221\tiny{$\pm0.005$} &    0.238 &  0.240\tiny{$\pm0.001$} &          0.249 \\
$\mathcal{F}_{targeted}$ & BA & 2  &  0.010\tiny{$\pm0.001$} & 0.022 & 0.018 & 0.018 &   0.045 &  0.022\tiny{$\pm0.002$} &    0.033 &  0.042\tiny{$\pm0.001$} &          \textbf{0.047} \\
                 &                & 5  &  0.025\tiny{$\pm0.001$} & 0.091 & 0.037 & 0.077 &   0.077 &  0.055\tiny{$\pm0.003$} &    0.077 &  0.108\tiny{$\pm0.001$} &          \textbf{0.117} \\
                 &                & 10 &  0.054\tiny{$\pm0.003$} & 0.246 & 0.148 & 0.232 &   0.116 &  0.128\tiny{$\pm0.014$} &    0.217 &  0.272\tiny{$\pm0.002$} &          \textbf{0.289} \\
                 & ER & 2  &  0.020\tiny{$\pm0.002$} & 0.103 & 0.090 & 0.098 &   \textbf{0.149} &  0.102\tiny{$\pm0.002$} &    0.118 &  0.122\tiny{$\pm0.001$} &          0.128 \\
                 &                & 5  &  0.050\tiny{$\pm0.002$} & 0.205 & 0.166 & 0.215 &   \textbf{0.293} &  0.182\tiny{$\pm0.008$} &    0.238 &  0.268\tiny{$\pm0.001$} &          0.279 \\
                 &                & 10 &  0.098\tiny{$\pm0.003$} & 0.306 & 0.274 & 0.299 &   0.477 &  0.269\tiny{$\pm0.016$} &    0.374 &  0.461\tiny{$\pm0.003$} &          \textbf{0.482} \\
\end{tabular}
}

\end{small}
\caption{Mean cumulative reward per episode obtained by the agents on synthetic graphs with $|V|=20$, grouped by objective function, graph family, and number of edge additions $L$. Each reported value represents the improvement in the expected critical fraction between the final and initial graphs.}
\label{tab:main_results}
\end{table*}

\begin{figure*}[t]
\centering
\includegraphics[width=0.99\textwidth]{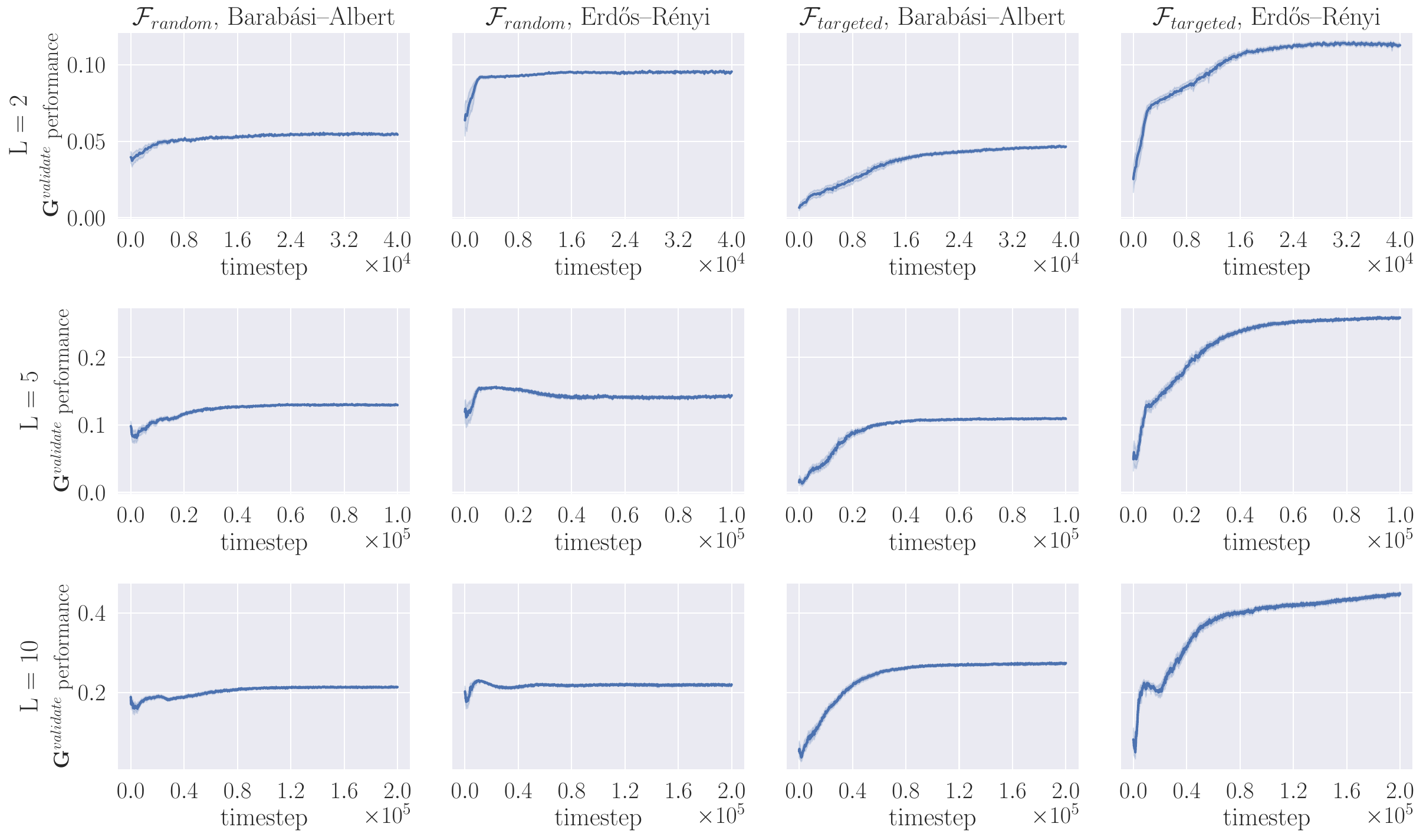} 
\caption{Performance on $\mathbf{G}^{validate}$ for synthetic graphs as a function of training steps. \customhl{Note the different x-axes scales for each row: more training steps are typically required for longer edge addition sequences.}}
\label{fig_eval_curves}
\end{figure*}

\begin{figure*}[t]
\centering
\includegraphics[width=0.99\textwidth]{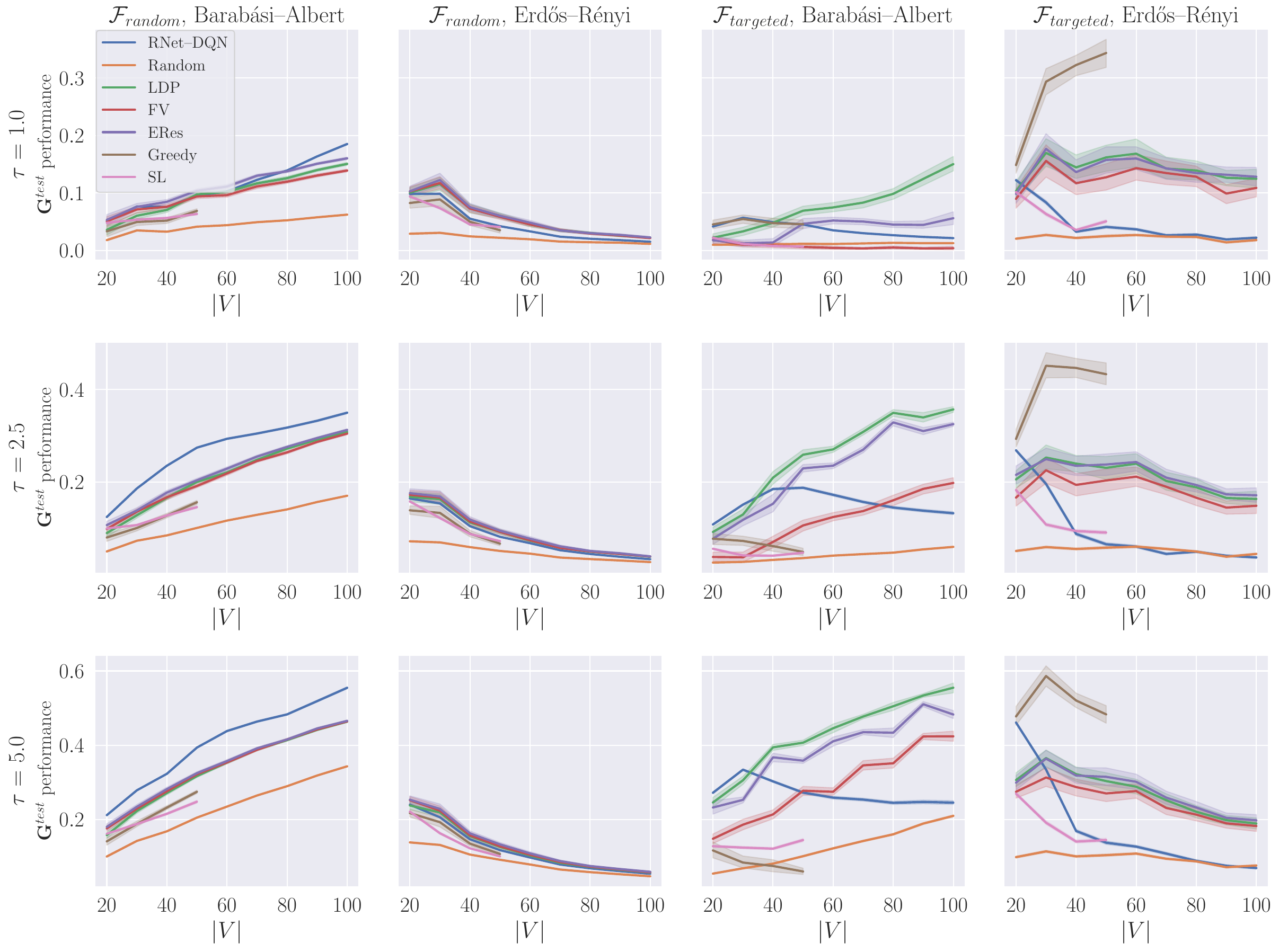} 
\caption{Performance on out-of-distribution synthetic graphs as a function of graph size, grouped by target problem and percentage of edge additions $\tau$. For \algoname and SL, models trained on graphs with $|V|=20$ are used.}
\label{fig_size_perf}
\end{figure*}

\begin{figure*}[ht!]
\centering
\includegraphics[width=0.99\textwidth]{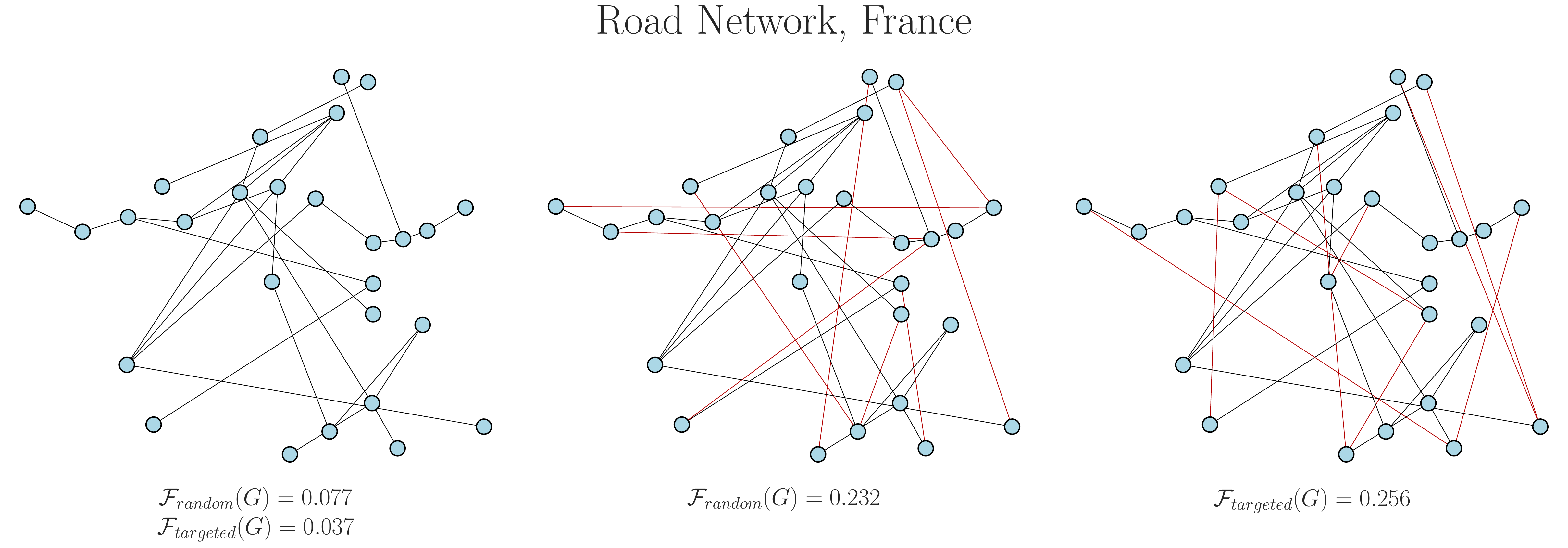}\vspace{0.01\textheight}
\includegraphics[width=0.99\textwidth]{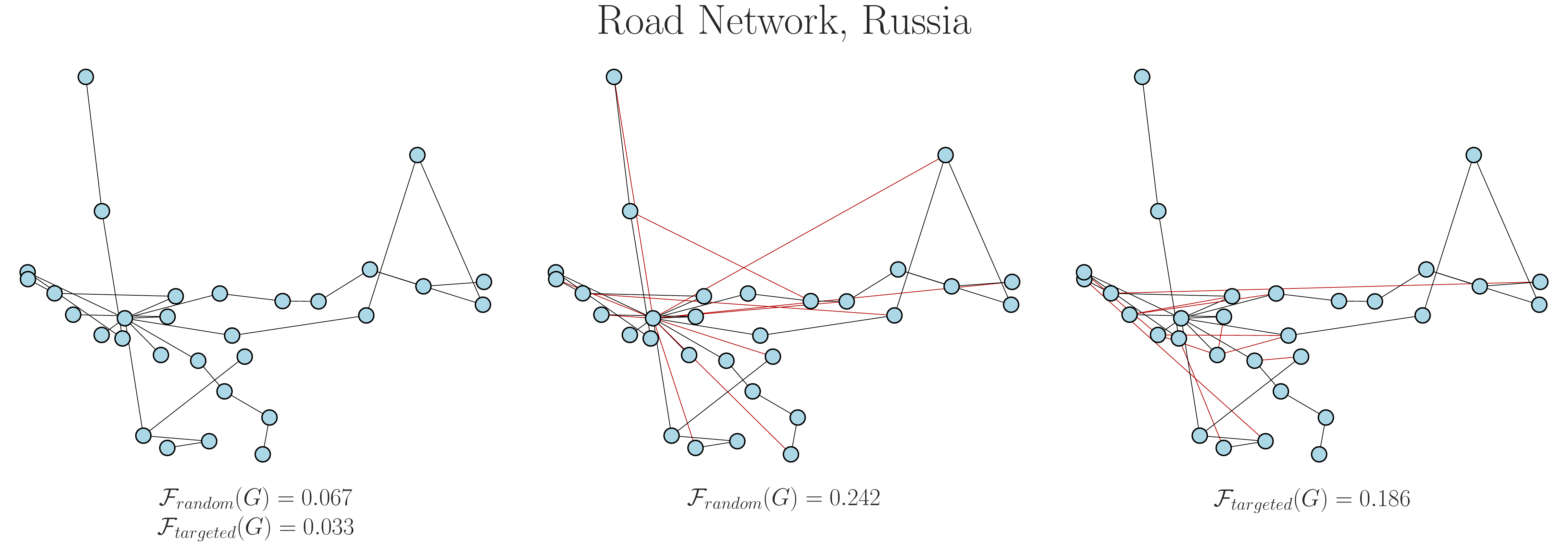}\vspace{0.01\textheight}
\includegraphics[width=0.99\textwidth]{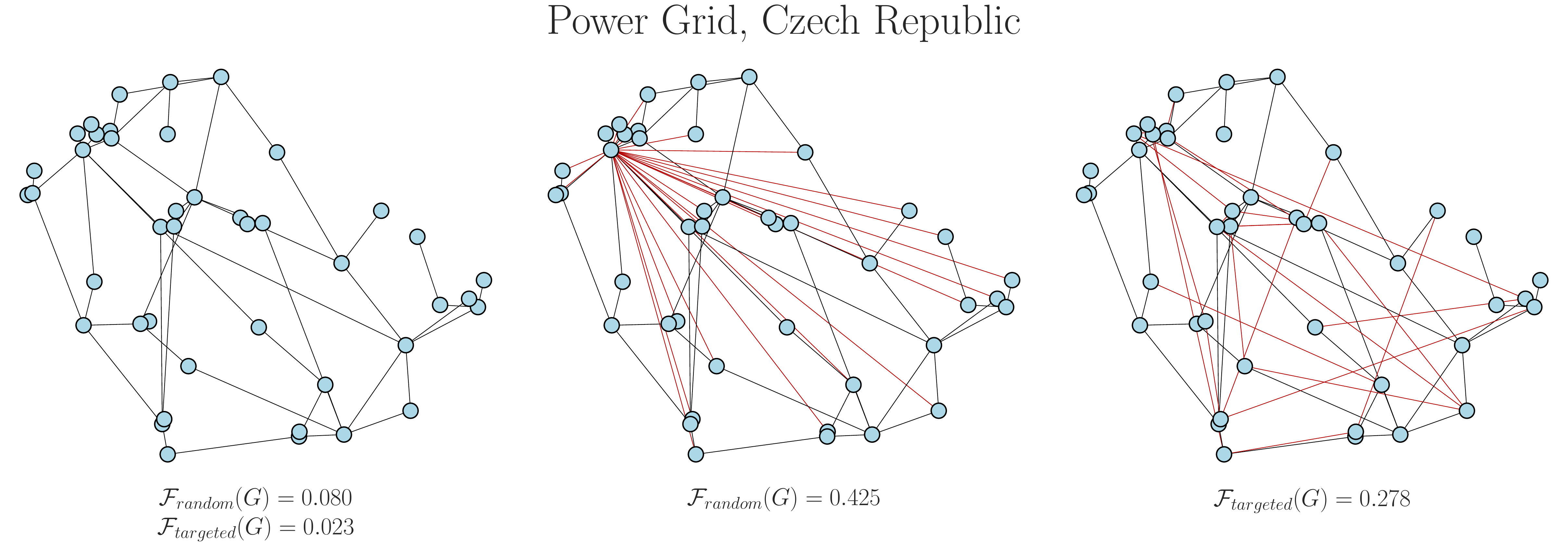}\vspace{0.01\textheight}
\caption{Several examples of the solutions found by \algoname on real-world graphs. Each row of the illustration shows the original network on the left, while the central and right panels show the network optimised for resilience to random and targeted removals, respectively. Objective function values are shown underneath. \customhl{The solutions for $\mathcal{F}_{random}$ typically assign more connections to a few central nodes, notably discovering the hub pattern in the third example. For $\mathcal{F}_{targeted}$ the added edges are spread around the network, reducing the impact of attacks. However, the algorithm might discover more complex patterns that are not directly interpretable, as shown in the solutions for the first example network.}}
\label{fig_rw_examples}
\end{figure*}

In Table~\ref{tab:main_results}, we present the results of our experimental evaluation for synthetic graphs. We also display the evolution of the validation loss during training in Figure~\ref{fig_eval_curves}. Out-of-distribution generalisation results are shown in Figure~\ref{fig_size_perf}. The results for real-world graphs are provided in Table~\ref{tab:rw_extended}. Additionally, Figure~\ref{fig_rw_examples} displays examples of the original and improved topologies found by our approach.

\begin{table*}
\centering
\begin{small}

\resizebox{\columnwidth}{!}{
\begin{tabular}{ccl|ccccccc}
                 &         &  &       Random &   LDP &    FV &  ERes &  Greedy &     \multicolumn{2}{c}{\algoname}   \\
Objective & Dataset & Instance &              &       &       &       &         & \textit{avg}& \textit{best} \\
\midrule
$\mathcal{F}_{random}$ & Euroroad & Finland &  0.080\tiny{$\pm0.019$} & 0.133 & 0.163 & 0.170 &   0.162 &  0.161\tiny{$\pm0.015$} &          \textbf{0.189} \\
                 &         & France &  0.057\tiny{$\pm0.016$} & 0.149 & 0.181 & 0.163 &   0.151 &  0.178\tiny{$\pm0.013$} &          \textbf{0.202} \\
                 &         & Kazakhstan &  0.107\tiny{$\pm0.018$} & 0.165 & 0.191 & 0.180 &   0.160 &  0.179\tiny{$\pm0.010$} &          \textbf{0.203} \\
                 &         & Poland &  0.082\tiny{$\pm0.033$} & 0.186 & 0.170 & 0.201 &   0.140 &  0.196\tiny{$\pm0.014$} &          \textbf{0.230} \\
                 &         & Romania &  0.076\tiny{$\pm0.025$} & 0.196 & 0.170 & \textbf{0.243} &   0.203 &  0.207\tiny{$\pm0.013$} &          0.235 \\
                 &         & Russia &  0.084\tiny{$\pm0.016$} & 0.135 & 0.224 & 0.157 &   0.187 &  0.199\tiny{$\pm0.017$} &          \textbf{0.230} \\
                 &         & Turkey &  0.092\tiny{$\pm0.023$} & 0.191 & 0.198 & 0.198 &   0.191 &  0.215\tiny{$\pm0.011$} &          \textbf{0.247} \\
                 &         & Ukraine &  0.071\tiny{$\pm0.017$} & 0.158 & 0.186 & 0.151 &   0.098 &  0.163\tiny{$\pm0.022$} &          \textbf{0.205} \\
                 & Scigrid & Switzerland &  0.050\tiny{$\pm0.035$} & 0.191 & 0.160 & 0.174 &   0.182 &  0.198\tiny{$\pm0.017$} &          \textbf{0.226} \\
                 &         & Czech Republic &  0.091\tiny{$\pm0.020$} & 0.242 & 0.239 & 0.252 &   0.214 &  0.334\tiny{$\pm0.020$} &          \textbf{0.375} \\
                 &         & United Kingdom &  0.111\tiny{$\pm0.020$} & 0.263 & 0.273 & 0.290 &   0.224 &  0.321\tiny{$\pm0.022$} &          \textbf{0.379} \\
                 &         & Hungary &  0.051\tiny{$\pm0.029$} & 0.176 & 0.179 & 0.175 &   0.117 &  0.148\tiny{$\pm0.017$} &          \textbf{0.185} \\
                 &         & Ireland &  0.090\tiny{$\pm0.014$} & 0.208 & 0.211 & 0.213 &   0.177 &  0.201\tiny{$\pm0.013$} &          \textbf{0.228} \\
                 &         & Sweden &  0.097\tiny{$\pm0.029$} & 0.187 & 0.213 & 0.195 &   0.197 &  0.213\tiny{$\pm0.022$} &          \textbf{0.276} \\
$\mathcal{F}_{targeted}$ & Euroroad & Finland &  0.069\tiny{$\pm0.018$} & 0.149 & 0.112 & 0.112 &   \textbf{0.307} &  0.273\tiny{$\pm0.009$} &         0.300 \\
                 &         & France &  0.032\tiny{$\pm0.019$} & 0.199 & 0.120 & 0.120 &   0.074 &  0.218\tiny{$\pm0.006$} &          \textbf{0.228} \\
                 &         & Kazakhstan &  0.052\tiny{$\pm0.021$} & 0.161 & 0.137 & 0.124 &   0.229 &  0.236\tiny{$\pm0.014$} &          \textbf{0.257} \\
                 &         & Poland &  0.010\tiny{$\pm0.008$} & 0.101 & 0.114 & 0.084 &   0.108 &  0.230\tiny{$\pm0.008$} &          \textbf{0.248} \\
                 &         & Romania &  0.029\tiny{$\pm0.021$} & 0.167 & 0.056 & 0.126 &   0.148 &  0.238\tiny{$\pm0.021$} &          \textbf{0.270} \\
                 &         & Russia &  0.000\tiny{$\pm0.000$} & 0.000 & 0.000 & 0.053 &   0.000 &  0.110\tiny{$\pm0.036$} &          \textbf{0.155} \\
                 &         & Turkey &  0.044\tiny{$\pm0.021$} & 0.126 & 0.155 & 0.126 &   0.143 &  0.233\tiny{$\pm0.018$} &          \textbf{0.264} \\
                 &         & Ukraine &  0.031\tiny{$\pm0.023$} & 0.074 & 0.037 & 0.083 &   0.135 &  0.164\tiny{$\pm0.006$} &          \textbf{0.178} \\
                 & Scigrid & Switzerland &  0.030\tiny{$\pm0.024$} & 0.000 & 0.103 & 0.098 &   0.045 &  0.128\tiny{$\pm0.006$} &          \textbf{0.139} \\
                 &         & Czech Republic &  0.038\tiny{$\pm0.026$} & 0.116 & 0.116 & 0.116 &   0.163 &  0.242\tiny{$\pm0.027$} &          \textbf{0.284} \\
                 &         & United Kingdom &  0.070\tiny{$\pm0.047$} & 0.190 & 0.095 & 0.184 &   0.207 &  0.252\tiny{$\pm0.027$} &          \textbf{0.326} \\
                 &         & Hungary &  0.027\tiny{$\pm0.028$} & \textbf{0.190} & 0.000 & 0.129 &   0.143 &  \textbf{0.190}\tiny{$\pm0.000$} &          \textbf{0.190} \\
                 &         & Ireland &  0.047\tiny{$\pm0.023$} & 0.101 & 0.084 & 0.106 &   0.079 &  0.259\tiny{$\pm0.011$} &          \textbf{0.288} \\
                 &         & Sweden &  0.061\tiny{$\pm0.021$} & 0.142 & 0.121 & 0.201 &   0.094 &  0.232\tiny{$\pm0.008$} &          \textbf{0.261} \\
\end{tabular}
}

\end{small}
\caption{Results obtained on real-world graphs, split by graph instance. Each reported value represents the improvement in the expected critical fraction between the final and initial graphs.}
\label{tab:rw_extended}
\end{table*}

\paragraph{\textbf{Main Findings.}} We summarise our findings as follows:

\noindent \textit{\algoname provides competitive performance, especially for longer action sequences.} Across all settings tested, \algoname performed significantly better than random. On synthetic graphs, the best model obtained the highest performance in 8 out of 12 settings tested, while the average performance is at least 89\% of that of the best-performing configuration. For BA graphs, \algoname obtained the best performance across all tasks tested. For ER graphs, ERes performed slightly better when considering $\mathcal{F}_{random}$; for $\mathcal{F}_{targeted}$ the greedy baseline performed better for shorter sequences. For real-world graphs, \algoname obtained the best performance across all tasks.

\noindent \textit{Strategies for improving $\mathcal{F}_{random}$ are easier to learn.} The performance gap between the trained model and the baselines is smaller for $\mathcal{F}_{random}$, suggesting it is less complex to learn. This is also supported by the evaluation losses monitored during training, which show performance improves and plateaus more quickly. For $\mathcal{F}_{random}$ the network with randomly initialised parameters already yields policies with satisfactory results, and training brings a small improvement. In contrast, the improvements for $\mathcal{F}_{targeted}$ are much more dramatic.

\noindent \textit{Out-of-distribution generalisation only occurs for $\mathcal{F}_{random}$.}
The performance on larger out-of-distribution graphs is preserved for the $\mathcal{F}_{random}$ objective, and especially for BA graphs we observe strong generalisation. The performance for $\mathcal{F}_{targeted}$ decays rapidly, obtaining worse performance than the baselines as the size increases. The poor performance of the greedy policy means the $Q(s,a)$ estimates are no longer accurate under distribution shift. There are several possible explanations, e.g., the inherent noise of estimating $\mathcal{F}_{random}$ makes the neural network more robust to outliers, or that central nodes impact message passing in larger graphs differently. We think investigating this phenomenon is a worthwhile future direction of this work, since out-of-distribution generalisation does occur for $\mathcal{F}_{random}$ and evaluating the objective functions directly is prohibitively expensive for large graphs.

\noindent \textit{Performance on real-world graphs is comparatively better wrt. the baselines.} This is expected since training is performed separately for each graph to be optimised.

\noindent \paragraph{Time Complexity.} 
We also compare the time complexities of all approaches considered below.
\begin{itemize}
\item \textit{\algoname}: $O(|V|+|E|)$ operations at each step: constructing node and graph-level embeddings and, based on these embeddings, performing the forward pass in the neural network to estimate $Q(s,a)$ for all valid actions.
\item \textit{Random}: $O(1)$ for sampling, assuming the environment checks action validity.
\item \textit{Greedy}: $O(|V|^4 \times (|V|+|E|))$. The improvement in $\mathcal{F}$ is estimated for all $O(|V|^2)$ possible edges. For each edge, this involves $O(|V|)$ MC simulations. As described in Section~\ref{problem}, each MC simulation has complexity $O(|V| \times (|V|+|E|))$.
\item \textit{LDP}: $O(|V|^2)$: computing the product of node degrees.
\item \textit{FV, ERes}: $O(|V|^3)$, since they involve computing the eigendecomposition and the Moore-Penrose pseudoinverse of the graph Laplacian respectively (may be faster in practice).  
\item \textit{SL}: $O(|V|^2\times(|V|+|E|))$. $\mathcal{\hat{F}}$ is predicted for $O(|V|^2)$ graphs that are one edge away, then an $argmax$ is taken.
\end{itemize}

It is worth noting that the analysis above does not account for the cost of training, the complexity of which is difficult to determine as it depends on many hyperparameters and the specific characteristics of the problem at hand. The approach is thus advantageous in situations in which predictions need to be made quickly, over many graphs, or the model transfers well from a cheaper training regime. We also remark that, even though the method requires an upfront cost for training, this can be seen as a constant term if the number of problem instances over which we would like to obtain predictions is large. These characteristics are shared with other emergent work that tackles combinatorial optimisation with machine learning~\cite{khalil_learning_2017,bengio_machine_2018}.

\section{Related Work and Discussion}\label{related}
\paragraph{Network Resilience.} Network resilience was first quantified by the average shortest path distance as a function of the number of removed nodes~\cite{albert_error_2000}. Analysing two scale-free communication networks, the authors found that this type of network has good robustness to random failure but is vulnerable to targeted attacks. A more extensive investigation~\cite{holme_attack_2002} analysed the robustness of several real-world networks as well as some generated by synthetic models using a variety of attack strategies. Another area of interest is the analysis of the phase transitions of the graph in terms of connectivity under the two attack strategies \cite{cohen_resilience_2000,cohen_breakdown_2001}. Optimal network topologies have also been discovered -- for example, under the objective of resilience to both failures and attacks, the optimal network has a bi-modal or tri-modal degree distribution~\cite{valente_two-peak_2004,tanizawa_optimization_2005}. There exists evidence to suggest that the topological robustness of infrastructure systems is correlated to operational robustness~\cite{soleRobustnessEuropeanPower2008}. More broadly, the resilience of systems is highly important in structural engineering and risk management~\cite{cimellaroFrameworkAnalyticalQuantification2010,ganinOperationalResilienceConcepts2016}.

\paragraph{Graph Neural Networks and Combinatorial Optimisation.} Neural network architectures able to deal not solely with Euclidean but also with manifold and graph data have been developed in recent years~\cite{bronstein_geometric_2017}, and applied to a variety of problems where their capacity for representing structured and relational information can be exploited~\cite{battaglia_relational_2018}. A sub-category of such approaches are Message Passing Neural Networks (MPNN)~\cite{gilmer_neural_2017}, often referred to as Graph Neural Networks (GNN) instead. Significant progress has been achieved in machine learning for combinatorial optimisation problems~\cite{bengio_machine_2018}, such as Minimum Vertex Cover and the Travelling Salesman Problem by framing them as a supervised learning~\cite{vinyals_pointer_2015} or RL~\cite{bello_neural_2016} task. Combining GNNs with RL algorithms has yielded models capable of solving several graph optimisation problems with the same architecture while generalising to graphs an order of magnitude larger than those used during training~\cite{khalil_learning_2017}. The problem as formulated in this paper is a combinatorial optimisation problem, related to problems in network design that arise in operations research~\cite{ahujaChapterApplicationsNetwork1995} -- albeit with a different objective.

\paragraph{Graph Generation.} Similarities exist between the current work and the area of graph generative modelling, which tries to learn a generative model of graphs~\cite{li_learning_2018,you_graphrnn_2018,liao_efficient_2019} in a computationally efficient way given some existing examples. This generation, however, is not necessarily conditioned by an objective that captures a global property to be optimised. Other lines of work target constructing, \textit{ab initio}, graphs that have desirable properties: examples include neural architecture search~\cite{zoph_neural_2016,liu_hierarchical_2018} and molecular graph generation~\cite{jin_junction_2019,you_graph_2018,bradshawModelSearchSynthesizable2019}. The concurrent work GraphOpt~\cite{trivedi_graphoptlearning_2020} tackles the \textit{inverse problem}: given a graph, the goal is to learn the underlying objective function that leads to its generation. This is achieved with maximum entropy inverse RL and GNNs. 

\paragraph{Relationship to RL--S2V.} Our work builds on RL--S2V, a method that was applied for the construction of adversarial examples against graph classifiers~\cite{dai_adversarial_2018}. However, it is worth noting that there are a series of key differences with respect to that approach. First, RL--S2V is not designed to address the problem of constructing robust graphs or, more generally, learning to construct graphs according to a given goal. Secondly, there are two key algorithmic differences to RL--S2V  with respect to the \mdpshort formulation: the reward function used, which in this case quantifies a global property of the graph itself, as well as the definition of the action spaces and the transition model, which account for excluding already-existing edges (RL--S2V ignores this, leading to some of the edge budget being wasted). Since the values of structural properties we consider are increasing in the number of edges (the complete graph has robustness 1), \algoname generally yields strictly better performance results.
\section{Conclusion and Outlook}\label{conclusion}
In this work, we have addressed the problem of improving a graph structure given the goal of maximising the value of a global objective function. We have framed it for the first time as a decision-making problem and we have formalised it as the \mdplong (\mdpshort). Our approach, named \textit{\algoname}, uses Reinforcement Learning and Graph Neural Networks as key components for generalisation. As a case study, we have considered the problem of improving graph robustness to random and targeted removals of nodes. Our experimental evaluation on synthetic and real-world graphs shows that, in certain situations, this approach can deliver performance superior to existing methods, both in terms of the solutions found (i.e., the resulting robustness of the graphs) and time complexity of model evaluation. Further, we have shown the ability to transfer to out-of-sample graphs, as well as the potential to transfer to out-of-distribution graphs larger than those used during training.

\paragraph{Extensions.} The proposed approach can be applied to other problems based on different definitions of robustness or considering fundamentally different objective functions such as efficiency~\cite{latoraEfficientBehaviorSmallWorld2001}, path diversity~\cite{gvozdiev_lowlatencycapabletopologies_2018}, and assortativity~\cite{newman_assortative_2002}, which are of interest in various biological, communication, and social networks. Since our formulation and algorithm are objective-agnostic, we expect they are applicable out-of-the-box for other objectives, even those for which no strong baselines are currently known. As such, this approach may be a useful tool for the discovery of new graph improvement algorithms \customhl{for objectives that can be evaluated programatically, either in closed form or via simulations. Potential limitations might be related to the complexity of the objective functions and the related computational demands. We also view the interpretability of the approach, which is less straightforward than those based on known mathematical concepts such as the Fiedler vector, to be an important research direction. Since there is an active interest in the interpretability of both GNNs and RL~\mbox{\cite{ying2019gnnexplainer,verma2018programmatically}}, we consider that there is scope for developing techniques that are tailor-made for explaining policies learned by RL on graphs.}

\paragraph{Operationalisation.} \customhl{Beyond considering other objectives, in order to operationalise the proposed algorithm, it is possible to integrate a variety of refinements, which can include capturing heterogeneous edge costs (e.g., different capacities per link in a communication network), extending the action space to support heterogenous edge types (e.g., addition of different types of edges with specific characteristics), and integrating domain-specific link constraints (e.g., planarity). For critical scenarios, it is also possible to verify that the resulting solutions satisfy some given formal properties and constraints~\mbox{\cite{garciaComprehensiveSurveySafe2015}}. Furthermore, considering multi-criteria objective functions~\mbox{\cite{roijers_surveymultiobjective_2013}} is important for cases where properties of the solutions must be balanced~\mbox{\cite{keeney1993decisions}}. Various choices exist for representing this trade-off and should be captured on a case-by-case basis depending on the application: for example, a linear combination may be sufficient in certain situations, while others are characterised by economies of scale. We also remark that our formulation captures operational scenarios in which the cost of constructing a link is significantly greater than the cost of its maintenance (e.g., as with road networks). Our method can also be adapted for situations in which operational cost is instead greater by formulating a rewiring operation that keeps the number of links constant.}

\paragraph{Broader Applications.} The approach described in this work can be used to improve the properties of a variety of human-made infrastructure systems, such as communication networks, transportation networks, and power grids. We also envisage potential applications in biological (e.g., hypothesis testing for understanding the characteristics of brain networks~\cite{bullmore_economy_2012}), ecological (e.g., design of more resilient ecosystems~\cite{westman1978measuring}) and social networks (e.g., design of organisational structures~\cite{wreathall2017properties}). 
As far as biological networks are concerned, the brain is hypothesised to optimise a trade-off between efficiency and wiring cost~\cite{bullmore_economy_2012}. Our method could be used in order to test different hypotheses related to the resulting structure of brain networks over evolutionary times and also during their development. Indeed, the evolution of such networks in time has been captured (for example, Sulston et al.~\cite{sulston_embryonic_1983} mapped the development of the C. elegans connectome). This can be achieved by applying the optimisation procedure for different objective functions and comparing the obtained networks to the ``ground truth''.
With respect to the potential application in ecosystem management, the graph formalisation can be used to model interactions between species in a given environment. As such, our method has potential applications to study, in simulation, the impact of introducing or removing species from an ecosystem so as to achieve a desired outcome. Specifically, in the context of robustness, we can consider optimising the resilience of an ecosystem to intrinsic or extrinsic shocks, a task of fundamental importance~\cite{allesina_functional_2009}. The dynamics of interactions between species may be modelled in simulation using well known models of e.g., predator-prey mechanics~\cite{berryman_origins_1992}.
With respect to social networks, for example, our method can be applied to derive optimal communication strategies and related team structures so as to optimise a given objective for an organisation. 
Finally, there are also potential applications for networks of artificial agents (i.e., robots). There is a significant body of work in the robotics literature that treats the problem of maintaining robust communication in a network of agents working together to complete a task in an environment that contains obstacles or adversaries. For instance,~\cite{stump_connectivity_2008} uses properties of the graph Laplacian (namely, the Fiedler vector and its associated eigenvalue) to ensure the underlying communication network remains robust. Since we have empirically shown superior performance to using the Fiedler vector, our approach could also lead to gains in this deployment scenario.
\vskip6pt

\enlargethispage{20pt}

\paragraph{Data access.} The original real-world datasets used in this research (Scigrid, Euroroad) are publicly available and were retrieved via the Scigrid project website \url{https://www.power.scigrid.de/pages/downloads.html} and KONECT \url{http://konect.cc/} respectively. They can be downloaded from their respective portals without registration. Scigrid is licensed under the Open Database License (ODbL) v1.0, while Euroroad is license-free. The scripts and instructions used to extract the subgraphs corresponding to individual countries in the infrastructure networks are \customhl{available in the code repository, which is part of the Supplementary Material.}

\paragraph{Author contributions.} V.-A.D., S.H., M.M. designed and developed the study, reviewed the results, and wrote the manuscript. V.-A.D. wrote the implementation and performed the data analysis. All authors gave final approval for publication and agree to be held accountable for the work performed therein.

\paragraph{Competing interests.} The authors declare no competing interests with respect to this work.

\paragraph{Funding.} This work was supported by The Alan Turing Institute under the UK EPSRC grant EP/N510129/1.

\vskip2pc

\bibliographystyle{plain}

\bibliography{bibliography} %

\newpage
\newpage
\section*{Supplementary Material}

\subsection*{Implementation Details}\label{impl}

Source code is available at~\href{https://rs.figshare.com/articles/dataset/Source_Code_from_Goal-directed_graph_construction_using_reinforcement_learning/16843164}{this URL}. For full details about how to set up the experimental infrastructure, run the experiments, and reproduce the results please see the~\href{https://rs.figshare.com/articles/journal_contribution/Code_Instructions_from_Goal-directed_graph_construction_using_reinforcement_learning/16843167}{instructions}.

The \algoname implementation uses PyTorch and is bootstrapped from the RL-S2V implementation provided by Dai et al.\footnote{\url{https://github.com/Hanjun-Dai/graph_adversarial_attack}}, which is based on the authors' implementation of the S2V GNN\footnote{\url{https://github.com/Hanjun-Dai/pytorch_structure2vec}}. We implement the performance-critical robustness simulations in a custom C++ module. Details of how the robustness objective functions are calculated are shown in Algorithm 1.

\begin{algorithm}[]
\caption{Estimating the robustness of graphs to targeted and random removals of nodes.}
\label{alg:robustness_estimation}
\textbf{Input}: undirected graph $G$ \\
\textbf{Parameters}: node removal strategy $S$, number of Monte Carlo simulations $K$ \\
\textbf{Output}: estimated robustness $\mathcal{F}$
\begin{algorithmic}[1] %
\State $r =$ \Call{array}{$K$}
\State $n =$ \Call{nodes}{$G$}
\State $N =$ \Call{len}{$n$}
\For{simulation $i=1$ to $K$ } \Comment{Trivially parallelisable}
    \State $p\gets$ \Call{generate\_permutation}{$G,S$} \Comment{Depends on $S$}
    \For{sequence index $j=1$ to $N$ }
        \State $m\gets n[p[j]]$ \Comment{Select next node in permutation}
        \State \Call{remove\_node}{$G,m$}  \Comment{Remove the node and all its edges}
        \State $c\gets$ \Call{num\_connected\_components}{$G$}
        \If{$c > 1$} 
            \State $f\gets j/N$ \Comment{Critical fraction reached}
            \State $r[i]\gets f$ 
            \State break 
        \EndIf
    \EndFor
\EndFor
\State \textbf{return} \Call{mean}{$r$}

\end{algorithmic}
\end{algorithm}

\subsection*{Hyperparameter Details}\label{hyperparameter}
\paragraph{General Parameters.}
We train for $4*10^4 \tau$ steps using the Adam optimiser and a batch size of $50$. For \algoname, we use an experience replay buffer of size equal to the number of steps. We let $\gamma=1$, since we are in the finite horizon case. Target network weights are updated every $50$ steps. For SL, we perform early stopping if the validation loss does not improve after $10^4$ steps. We do not use any weight regularisation or normalisation. We do not perform any gradient clipping when computing the DQN or MSE losses. Weights are initialised using Glorot initialisation. We use a learning rate $\alpha=0.0001$. During training, we scale the rewards linearly by a factor of $100$ in order to improve numerical stability. 

\paragraph{Parameters for Synthetic Graphs.} For synthetic graphs, we use a number of message passing rounds $K=3$. The MLP layer has $128$ hidden units. For \algoname, we decay the exploration rate $\epsilon$ linearly from $\epsilon=1$ to $\epsilon=0.1$ for the first half of training steps, then fix $\epsilon=0.1$ for the rest of the training. To estimate the values of the objective functions we use $2|V|$ Monte Carlo simulations.

\paragraph{Parameters for Real-World Graphs.} For real-world infrastructure graphs we use $K=5$ as they are larger in size and diameter. The MLP layer has $32$ hidden units. For \algoname, we decay $\epsilon$ linearly from $\epsilon=1$ to $\epsilon=0.1$ in the first 10\% of training steps, then fix $\epsilon=0.1$ for the rest of the training. To estimate the values of the objective functions we use $40$ Monte Carlo simulations. 

\subsection*{Runtime Details}\label{runtime}
For the computational experiments presented in this paper, we used a machine with 2 Intel Xeon E5-2637 v4 processors, 64GB RAM, and a NVIDIA Tesla P100 GPU. For synthetic graphs, the time per \algoname training run for a specific robustness objective function and graph family is approximately 3.5 hours with $L=10$. The computational time for the synthetic graph experiments is approximately 1350 hours (56 days), while the experiments on real-world graphs took approximately 1240 hours (51 days). Since most of the cost is due to considering a large number of random initialisations in order to provide statistically robust evaluations, the experiments can be trivially parallelised.  

\end{document}